\documentclass[journal]{IEEEtran}

\ifCLASSINFOpdf
\else
   \usepackage[dvips]{graphicx}
\fi
\usepackage{url}

\usepackage{graphicx}
\usepackage{times}
\usepackage{epsfig}
\usepackage{graphicx}
\usepackage{amsmath,amssymb,amsfonts}
\usepackage{algorithmic}
\usepackage{textcomp}
\usepackage{subfigure}
\usepackage{caption}
\usepackage{color}
\usepackage{xcolor}
\usepackage{booktabs}
\usepackage{arydshln}
\usepackage{bbding}
\usepackage{multirow}
\usepackage{diagbox}
\usepackage{mathrsfs}
\usepackage{makecell}
\usepackage{enumitem}

\usepackage{colortbl}
\usepackage{arydshln}
\usepackage{wrapfig}
\usepackage{sidecap}
\usepackage{balance}
\usepackage[pagebackref=false,breaklinks=true,letterpaper=true,colorlinks,bookmarks=false]{hyperref}
\usepackage{xspace}

\newcommand{\figref}[1]{Fig.~\ref{#1}}
\newcommand{\equref}[1]{Eq.~\eqref{#1}}
\newcommand{\secref}[1]{Sec.~\ref{#1}}
\newcommand{\tabref}[1]{Table~\ref{#1}}

\makeatletter
\DeclareRobustCommand\onedot{\futurelet\@let@token\@onedot}
\def\@onedot{\ifx\@let@token.\else.\null\fi\xspace}
\def\eg{\emph{e.g}\onedot} 
\def\ie{\emph{i.e}\onedot} 
 
\def\etc{\emph{etc}\onedot} 
\def\wrt{{w.r.t}\onedot}

\makeatother

\usepackage[font=small]{caption}
\makeatletter
\newcommand\figcaption{\def\@captype{figure}\caption}
\newcommand\tabcaption{\def\@captype{table}\caption}
\makeatother

\newcommand{\PreserveBackslash}[1]{\let\temp=\\#1\let\\=\temp}
\newcolumntype{C}[1]{>{\PreserveBackslash\centering}p{#1}}
\newcolumntype{R}[1]{>{\PreserveBackslash\raggedleft}p{#1}}
\newcolumntype{L}[1]{>{\PreserveBackslash\raggedright}p{#1}}

\begin{document}

\title{A Representation Separation Perspective to \\ Correspondences-free Unsupervised \\ 3D Point Cloud Registration}

\author{Zhiyuan Zhang$^1$, Jiadai Sun$^1$, Yuchao Dai$^1$ \IEEEmembership{Member, IEEE}, Dingfu Zhou$^{2}$,\\ Xibin Song$^{2}$, Mingyi He$^{1}$ \IEEEmembership{Senior Member, IEEE}
\thanks{This work was supported in part by the National Key Research and Development Program of China under Grant 2018AAA0102803 and the National Natural Science Foundation of China under Grants 61871325, 62001394 and 61901387.}
\thanks{Corresponding author: Yuchao Dai (e-mail: daiyuchao@nwpu.edu.cn).}
\thanks{$^1$School of Electronics and Information, Northwestern Polytechnical University, Xi'an, China. $^2$Baidu Research and National Engineering Laboratory of Deep Learning Technology and Application, Beijing, China.}
}

\markboth{Journal of \LaTeX\ Class Files, Vol. 14, No. 8, August 2015}
{Shell \MakeLowercase{\textit{et al.}}: Bare Demo of IEEEtran.cls for IEEE Journals}
\maketitle

\begin{abstract}
3D point cloud registration in remote sensing field has been greatly advanced by deep learning based methods, where the rigid transformation is either directly regressed from the two point clouds (correspondences-free approaches) or computed from the learned correspondences (correspondences-based approaches). Existing correspondences-free methods generally learn the holistic representation of the entire point cloud, which is fragile for partial and noisy point clouds.
In this paper, we propose a correspondences-free \underline{u}nsupervised \underline{p}oint \underline{c}loud \underline{r}egistration (\emph{UPCR}) method from the representation separation perspective. First, we model the input point cloud as a combination of pose-invariant representation and pose-related representation. Second, the pose-related representation is used to learn the relative pose \wrt a ``latent canonical shape'' for the \textit{source} and \textit{target} point clouds respectively. Third, the rigid transformation is obtained from the above two learned relative poses. Our method not only filters out the disturbance in pose-invariant representation but also is robust to partial-to-partial point clouds or noise.
Experiments on benchmark datasets demonstrate that our \textbf{unsupervised} method achieves comparable if not better performance than state-of-the-art \textbf{supervised} registration methods. 
\end{abstract}


\IEEEpeerreviewmaketitle

%%%%%%%%% BODY TEXT
%%%%%%%%%%%%%%%%%%%%%%%%%%%%%%%%%%%%%%%%%%%%%%%%%%%%%%%%%%%%
\section{Introduction} \label{sec:introduction}
%%%%%%%%%%%%%%%%%%%%%%%%%%%%%%%%%%%%%%%%%%%%%%%%%%%%%%%%%%%%

\IEEEPARstart{R}ECENTLY, with the widespread application of LiDAR, point cloud has become an important data source for 3D remote sensing, which presents rich 3D spatial information efficiently \cite{rong_pcc_rsl_2020,yan2015urban}. Meanwhile, to obtain a larger perceptive field and ensure the integrity of the objects in point cloud data, 3D point cloud registration has been widely investigated. The 3D point cloud registration task aims to estimate a rigid motion to align two 3D point clouds, 
% which is a fundamental task in geometrical computer vision, 
which is a key component in many applications ranging from remote object recognition and segmentation \cite{zhao_alsc_rsl_2020,luo_seg_rsl_2021}, simultaneous localization and mapping (SLAM) \cite{ding_deepmapping_cvpr_2019}, to autonomous driving \cite{wan_robustlocalization_icra_2018}, \etc. 
Most of the current 3D point cloud registration methods are built upon correspondences, devoting to minimizing the registration error between correspondences \cite{pais_3dregnet_cvpr_2020,perfect_Gojcic_cvpr19,3dlocal_Deng_CVPR_19,DeepGMR_yuan_eccv_20}. These methods are usually solved through two steps: correspondences building and transformation estimation. However, these correspondences-based methods, whether hand-crafted or learning-based, often suffer from a fatal drawback because they cannot handle outliers well.

As an alternative, the correspondences-free methods solve the rigid motion by comparing the holistic representations of the two point clouds. In the seminal work PointNetLK \cite{aoki_ptlk_cvpr_2019}, PointNet \cite{charles_pointnet_cvpr_2017} is used as the global feature extractor, and a modified LK \cite{lk_lucas_ijcai_1981} algorithm is designed for motion estimation. PCRNet \cite{sarode_pcrnet_arxiv_2019} inherits this framework and solves the rigid motion through regression. Note that these supervised methods rely on extensive registration label data. Feature-metric \cite{huang_featuremetric_cvpr_2020} is another correspondences-free method, which can be trained in a semi-unsupervised or unsupervised manner. This method devotes to designing feature extractors from the reconstruction perspective but does not exhibit sufficient robustness.

Essentially, these correspondences-free methods \cite{aoki_ptlk_cvpr_2019,sarode_pcrnet_arxiv_2019,huang_featuremetric_cvpr_2020} can be understood as solving the relative pose consuming the difference in global representations of the \textit{source} and \textit{target} point clouds. 
However, this operation is inaccurate and unreasonable, especially for partial-to-partial cases, which are common in the remote sensing field. 
Intuitively, the global representation of a point cloud can be viewed as a combination of the pose-related part (\ie, rotation and translation) and the pose-invariant part (\eg, shape, geometry structure, \etc). 
Thus, using the global representation to regress the final relative pose is problematic.
To deal with cases such as partial-view and outliers, which are pose-invariant,
the resulting network has to work as a \textbf{filter} due to the asymmetry of the inputs and outputs, thus leading to inaccurate performance.

To solve this problem, we propose a correspondences-free unsupervised 3D point cloud registration (\emph{UPCR}) method to learn the relative pose only from the pose-related part rather than from the global representation.
%%%   where the network works as a \textbf{transformer} because the input and output are equal. 
Specifically, \emph{UPCR} consists of three steps. First, we separate the input point cloud to a pose-invariant representation and a pose-related representation. Second, the pose-related representation is used to learn a relative pose \wrt a ``latent canonical shape'' for the \textit{source} and \textit{target} point clouds correspondingly. Finally, the 3D rigid transformation is obtained from the above two learned relative poses. In this way, the disturbance in pose-invariant representation is filtered out, which enables the high robustness of our method to partial point cloud and outliers.
We train our model in an unsupervised learning manner. Experiments on benchmark datasets demonstrate that our \textbf{unsupervised} method achieves comparable if not better performance than supervised 3D point cloud registration methods.

Our main contributions can be summarized as follows: 1) We present a correspondences-free unsupervised point cloud registration method from the representation separation perspective; 2) We design a representation separation module to decouple the pose-related representation from the global representation; 3) Experimental results on benchmark datasets validate the superiority and robustness of our method.

\vspace{-0.4cm}
%%%%%%%%%%%%%%%%%%%%%%%%%%%%%%%%%%%%%%%%%%%%%%%
\section{Proposed Approach}\label{sec:method}
%%%%%%%%%%%%%%%%%%%%%%%%%%%%%%%%%%%%%%%%%%%%%%%
\vspace{-0.0cm}
Given the \textit{source} and the \textit{target} point clouds $\mathbf{X} \in \mathbb{R}^{3\times N_\mathbf{X}}$, $\mathbf{Y} \in \mathbb{R}^{3\times N_\mathbf{Y}}$, the 3D registration problem aims at estimating the relative pose to achieve the best alignment, which can be modeled by a rotation matrix $\mathbf{R} \in SO(3)$ and a translation vector $\mathbf{t} \in \mathbb{R}^3$.  
We tackle this problem from a novel representation separation perspective. Specifically, we model the input point clouds as a combination of pose-invariant and pose-related representations. As the pose-invariant part does not provide motion estimation cues, we solve the relative pose only from the pose-related representation. The whole objective is formulated as:
\vspace{-7pt}
\begin{equation} 
    \mathop{\min}\limits_{\{\mathbf{T}_\mathbf{X},\mathbf{T}_\mathbf{Y}\}} \psi({\mathbf{R}_\mathbf{X}}^\mathrm{-1} (\mathbf{X} - \mathbf{t}_\mathbf{X}),{\mathbf{R}_\mathbf{Y}}^\mathrm{-1} (\mathbf{Y} - \mathbf{t}_\mathbf{Y})),% \space \mathbf{T}_\mathcal{X}\!=\!\Phi(\mathcal{X}), \mathbf{T}_\mathcal{Y}\!=\!\Phi(\mathcal{Y}),
\vspace{-4pt}
\end{equation}
where $\mathbf{T_{X}}=[\mathbf{R_{X}}|\mathbf{t_{X}}]$ and $\mathbf{T_{Y}}=[\mathbf{R_{Y}}|\mathbf{t_{Y}}]$ denote the relative poses learned only from the pose-related parts \wrt ``latent canonical shape'' $\mathbf{X}_\mathbf{c}$, $\mathbf{Y}_\mathbf{c}$, \ie $\mathbf{X} = \mathbf{R}_\mathbf{X} \mathbf{X}_\mathbf{c} + \mathbf{t}_\mathbf{X}$, $\mathbf{Y} = \mathbf{R}_\mathbf{Y} \mathbf{Y}_\mathbf{c} + \mathbf{t}_\mathbf{Y}$. Note that $\mathbf{X}_\mathbf{c}={\mathbf{R}_\mathbf{X}}^\mathrm{-1} (\mathbf{X} - \mathbf{t}_\mathbf{X})$ and $\mathbf{Y}_\mathbf{c}={\mathbf{R}_\mathbf{Y}}^\mathrm{-1} (\mathbf{Y} - \mathbf{t}_\mathbf{Y})$ are defined as the ``latent canonical shape'' with a ``canonical pose'', which are obtained by recovering the learned relative poses from the given shapes. Compared with setting a certain pose as the canonical pose artificially \cite{canonical_capsule_sun_2020_arxiv}, our canonical pose is learned in a data-driven manner. $\psi(\cdot,\cdot)$ is a metric function, which measures the difference of the input two canonical shapes. 
In this paper, we first summarize the pose-invariant representations of \textit{source} and \textit{target} denoted as $\Gamma_\nu^\mathbf{X}$, $\Gamma_\nu^\mathbf{Y}$, and the global representations as $\Gamma_\mathbf{G}^\mathbf{X}$, $\Gamma_\mathbf{G}^\mathbf{Y}$. Then a special subtraction is designed to solve the pose-related representations as $\Gamma_\mu^\mathbf{X}$, $\Gamma_\mu^\mathbf{Y}$ to regress respective relative pose $\mathbf{T_{X}}$, $\mathbf{T_{Y}}$ by subtracting pose-invariant representation from the global representation respectively. This special subtraction is introduced in detail below imitating the relative entropy.

%%%%%%%%%%%%%%%%%%%%%%%%%%%%%%%%%%%%%%%%%%%%%%%%%%%%%%%%%%%%%%%%%%%%%%%%%%%%%%%%%%%%%%%%%%
\vspace{-10pt}
\subsection{Representation Separation for Registration} \label{sec::representation_separation}
%%%%%%%%%%%%%%%%%%%%%%%%%%%%%%%%%%%%%%%%%%%%%%%%%%%%%%%%%%%%%%%%%%%%%%%%%%%%%%%%%%%%%%%%%%%
\vspace{-0.05cm}
Since we are modeling the point cloud as a combination of the pose-invariant part and the pose-related part, a natural idea is to remove the pose-invariant part from the global representation and what remains will be the pose-related part.

\textbf{Global Representation.} The global representation summarizes all information from the spatial coordinates of 3D points. Under the deep learning framework, many global representation extraction networks are proposed as standard modules \cite{LFNet_Cao_spl_21,ppf-foldnet_deng_eccv_2018,GANet_Deng_spl_21}. A pioneering work PointNet \cite{charles_pointnet_cvpr_2017} pools an arbitrary number of orderless points into a global descriptor. PointNet cannot summarize the local structure, while PointNet++ \cite{charles_pointnet2_nips_2017} alleviates this problem by a hierarchical structure. It is worth mentioning that DGCNN \cite{wang_dgcnn_tog_2019} encodes the local geometric characteristics by graph neural network (GNN) and achieves superior performance. Thus, we employ the GNN as our global representation extractor. 

Considering a point cloud $\mathbf{X}$, GNN constructs a K-nearest neighbor (KNN) graph, applies a non-linearity to the edge values, and performs vertex-wise aggregation (max-pooling) in each layer for point features, \ie,
\vspace{-0.1cm}
% \begin{equation}
\begin{gather}
\hspace{-2.52mm}
{\mathbf{F}^{\ell}(\boldsymbol{x}_i)} \!=\! f\!\left({\left\{ {h_\theta ^{\ell}\left({\mathbf{F}^{{\ell}-1}(\boldsymbol{x}_i)}, {\mathbf{F}^{{\ell}-1}(\boldsymbol{x}_{ij})} \right)|\!\ \forall {\boldsymbol{x}_{ij}} \!\in\! {\mathbb{N}({\boldsymbol{x}_i})}} \right\}} \right)\!,\!
% \raisetag{10pt}
\label{Eq:dgcnn_local}
% \end{equation}
\end{gather}
where $\mathbf{F}^{\ell}(\boldsymbol{x}_i)$ is the feature of point $\boldsymbol{x}_i$ in the $\ell$-th layer GNN, and $h_\theta ^{\ell}(\cdot)$ is a multi-layer perceptron (MLP) function parameterized by $\theta$. $f(\cdot)$ represents the symmetry function max-pooling, and $\mathbb{N}({\boldsymbol{x}_i})$ denotes the KNN neighbors of point ${\boldsymbol{x}_i}$, \ie\ ${\boldsymbol{x}_{ij}}$ is the $j$-th neighbor point, $j\in [1,K]$, $K$ is pre-defined. As a result, the global representation can be obtained by the max-pooling acting on all point features, $\Gamma_\mathbf{G}^{\mathbf{X}}=f(\{\mathbf{F}^{\ell}(\boldsymbol{x}_i)|i=1,\dots,N_{\mathbf{X}}\}) \in \mathbb{R}^{1\times m}$, $m$ is the pre-defined representation vector dimension. Similarly, $\Gamma_\mathbf{G}^{\mathbf{Y}} = f(\{\mathbf{F}^{\ell}(\boldsymbol{y}_i)|i=1,\dots,N_{\mathbf{Y}}\}) \in \mathbb{R}^{1\times m}$.

\textbf{Pose-Invariant Part Representation.} We extract the pose-invariant part representation of one point cloud via pose invariant feature. 
There are usually two methods to obtain invariant features, \ie, data augmentation and pose-invariant network. Data augmentation is easy to implement, but it is time-consuming. Thus, we adopt the second strategy to design a concise but effective feature called \textbf{distance-based feature}.

Considering a point cloud $\mathbf{X}$, whose center is denoted as ${\boldsymbol{o}}$, the distance-based point feature ${\phi}$ is designed as 
\vspace{-6pt}
\begin{equation}
{\phi} \left( {{{\boldsymbol{x}}_{ij}}} \right) \!=\! \left[ {D\left( {{{\boldsymbol{x}}_{ij}}, \boldsymbol{o}} \right),D\left( {{{\boldsymbol{x}}_{ij}},{{\boldsymbol{x}}_i}} \right),D\left( {\boldsymbol{o},{{\boldsymbol{x}}_i}} \right)} \right],
\vspace{-4pt}
\end{equation}
\noindent where ${\boldsymbol{x}}_{ij} \!\in\! \mathbb{N}({\mathbf{x}}_i)$ is a neighbor point as mentioned above. 
Function $D(\cdot,\cdot)$ calculates the Euclidean distance between two 3D points. $[\cdot,\cdot,\cdot]$ combines these three input elements to a point feature with 3 dimensions. Hence, ${\phi(\mathbf{x}_{ij})}$ maintains invariance to arbitrary rigid transformations. Then, using all neighbor points yields the pose-invariant feature of point ${\boldsymbol{x}}_i$,
\vspace{-6pt}
\begin{equation}
	\Phi^0({\boldsymbol{x}}_i) = f(\{h_{\alpha}(\phi({\boldsymbol{x}}_{ij}))\ |\ \forall{{\boldsymbol{x}}_{ij}} \in \mathbb{N}({\boldsymbol{x}}_i) \}),
\vspace{-4pt}
\end{equation}
\noindent where $h_{\alpha}$ is a MLP network parameterized by $\alpha$. According to \equref{Eq:dgcnn_local}, a GNN network is applied to purify the pose-invariant representation as: $\Gamma_\nu^\mathbf{X} = f(\{\Phi^{\ell}(\boldsymbol{x}_i)|i=1,\dots,N_{\mathbf{X}}\}) \in \mathbb{R}^{1\times m}$. Note that the dimension of pose-invariant part representation is same as the global representation. Similarly, $\Gamma_\nu^\mathbf{Y} = f(\{\Phi^{\ell}(\boldsymbol{y}_i)|i=1,\dots,N_{\mathbf{Y}}\}) \in \mathbb{R}^{1\times m}$.
Notably, while our distance-based feature characterizes pose-invariant feature, other pose-invariant feature can also play the same role within our pipeline. In \secref{sec:experiment}, we also report the performance of merging other advanced pose-invariant features including PFH \cite{PFH_rusu_iros_2008}, SPFH \cite{rusu_fpfh_icra_2009} into our UPCR method.
We find that, these more informative pose-invariant feature can generate more accurate registration results. And designed distance-based features helps to achieve advanced time-efficiency and registration performance simultaneously.

\begin{figure}[!t]
	\centerline{\includegraphics[width=9cm]{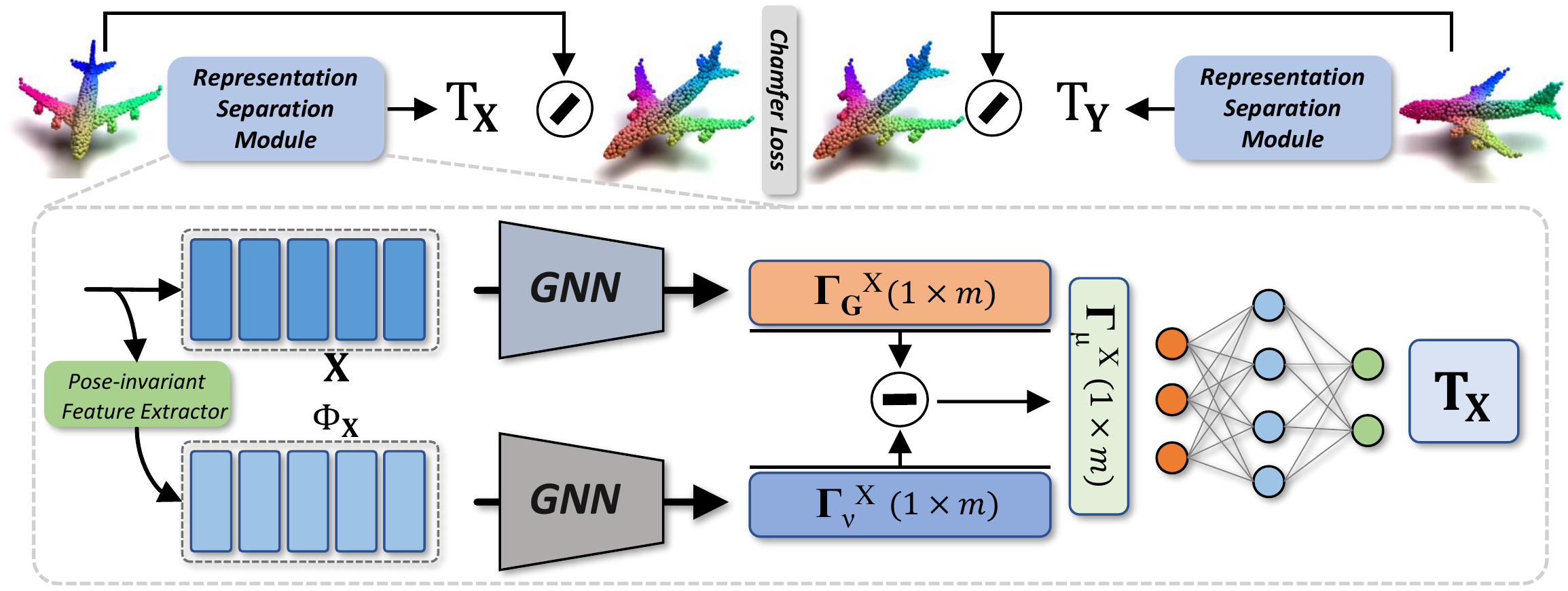}}
	\vspace{-0.1cm}
	\caption{Our UPCR network architecture. Given the \textit{source} and \textit{target} point clouds, the graph neural network is applied on the original point coordinates and the distance-based features to encode the global representation and pose-invariant part representation, respectively. Then, the pose-related part representation is produced by the representation separation module. Finally, we use it to regress their respective relative poses \wrt\ the same ``latent canonical shape''. And the final rigid transformation is obtained from these two learned poses.} 
	\label{Fig:network}
	\vspace{-0.5cm}
\end{figure}

\textbf{Pose-Related Part Representation.} Our core idea is to estimate the rigid motion by learning relative poses \wrt\ a ``latent canonical shape'' for \textit{source} and \textit{target} point clouds from pose-related part representations, which is more reasonable and accurate. To model this pose-related part representation, we propose to ``subtract'' pose-invariant part from global representation. Then, a natural question is, \textit{how to formulate this special subtraction}. A straightforward approach is to learn it through neural networks. However, this operation is uninterpretable and difficult to converge. In the following, we design a novel and elegant ``representation subtraction''.

Considering that the relative entropy often works as a ``distance'' between distributions, we formulate our subtraction analogously. Specifically, we first construct the probability distribution of the global representation and the pose-invariant representation, \ie\ $\mathbf{q} = \text{softmax}(\Gamma_\mathbf{G})$, $\mathbf{p} = \text{softmax}(\Gamma_\nu)$.
Then, we formulate the pose-related part representation as:
\vspace{-0.2cm}
\begin{equation}
\begin{aligned}
 \Gamma_\mu = [\mathbf{p}_{1}\log ({\mathbf{p}_{1}}/{\mathbf{q}_{1}}), \mathbf{p}_{2}\log ({\mathbf{p}_{2}}/{\mathbf{q}_{2}}), ..., \mathbf{p}_{m}\log ({\mathbf{p}_{m}}/{\mathbf{q}_{m}})],
\end{aligned}
\vspace{-0.1cm}
\end{equation}
where the subscript is the dimension index.

Then, we regress both rotations and translations from the pose-related representations. The Euler angle ${\Omega} \in \mathbb{R}^3$ is used to represent the rotation, which is compact and facilitates accurate regression. Afterwards, the relative poses $\mathbf{T}_\mathbf{X}$, $\mathbf{T}_\mathbf{Y}$ \wrt the same ``latent canonical shape'', which can be decoded as,
\vspace{-0.2cm}
\begin{equation}
\left\{ {\begin{aligned}
    {\Omega}_\mathbf{X}, \mathbf{t}_\mathbf{X}&=h_\beta (\Gamma_\mu^\mathbf{X}),\  \mathbf{R}_\mathbf{X} = \mathcal{T}({\Omega}_\mathbf{X}) \\
    {\Omega}_\mathbf{Y}, \mathbf{t}_\mathbf{Y}&=h_\beta (\Gamma_\mu^\mathbf{Y}),\ 
    \mathbf{R}_\mathbf{Y} = \mathcal{T}({\Omega}_\mathbf{Y})
\end{aligned}} \right.,
\vspace{-0.2cm}
\end{equation}
where $h_\beta$ is the MLP parameterized by $\beta$, $\mathcal{T}(\cdot)$ is the function that transforms Euler angle to rotation matrix. 
We obtain the final rigid transformation as ${\mathbf{R}} = \mathbf{R}_\mathbf{Y}  {\mathbf{R}_\mathbf{X}}^\mathrm{T}$, ${\mathbf{t}} = \mathbf{t}_\mathbf{Y} -  \mathbf{R}{\mathbf{t}_\mathbf{X}}$.

\vspace{-0.2cm}
%%%%%%%%%%%%%%%%%%%%%%%%%%%%%%%%%%%%%%%%%%%%%%%%%%%%%%%%%%%%%%%%%%%%%%%
\subsection{Unsupervised Loss}\label{sec::loss}
%%%%%%%%%%%%%%%%%%%%%%%%%%%%%%%%%%%%%%%%%%%%%%%%%%%%%%%%%%%%%%%%%%%%%%%
\vspace{-0.0cm}
Next, in our representation separation framework, we can transform the \textit{source} and \textit{target} point clouds into the canonical coordinate by $\mathbf{T}_\mathbf{X}$ and $\mathbf{T}_\mathbf{Y}$ respectively, \ie, $\mathbf{X}_\mathbf{c} = {\mathbf{R}_\mathbf{X}}^\mathrm{T} (\mathbf{X} - \mathbf{t}_\mathbf{X})$, $\mathbf{Y}_\mathbf{c} = {\mathbf{R}_\mathbf{Y}}^\mathrm{T} (\mathbf{Y} -\mathbf{t}_\mathbf{Y})$.
Note that the latent canonical shapes $\mathbf{X}_\mathbf{c}$ and $\mathbf{Y}_\mathbf{c}$ should be consistent. Here, we use the Chamfer distance to measure the discrepancy between $\mathbf{X}_\mathbf{c}$ and $\mathbf{Y}_\mathbf{c}$. In this case, any ground truth is not involved. In other words, an unsupervised loss is constructed as,  
\vspace{-0.2cm}
\begin{equation}
\begin{aligned}
Loss(\mathbf{X}_\mathbf{c},\mathbf{Y}_\mathbf{c}) =
 \frac{1}{N_{\mathbf{X}}} & \sum \limits_{\boldsymbol{x} \in \mathbf{X}_\mathbf{c}} \min \limits_{\boldsymbol{y}\in \mathbf{Y}_\mathbf{c}} \|\boldsymbol{x}-\boldsymbol{y}\|_2^2 \\
& + \frac{1}{N_{\mathbf{Y}}} \sum \limits_{\boldsymbol{y}\in \mathbf{Y}_\mathbf{c}} \min \limits_{\boldsymbol{x}\in \mathbf{X}_\mathbf{c}} \|\boldsymbol{x}-\boldsymbol{y}\|_2^2.
\end{aligned}
\vspace{-0.2cm}
\end{equation} 

%%%%%%%%%%%%%%%%%%%%%%%%%%%%%%%%%%%%%%%%%%%%%%%%%%%%%%%%%%%%%%%%%%%%%%%%%%%%%

%\vspace{-0.1cm}
%%%%%%%%%%%%%%%%%%%%%%%%%%%%%%%%%%%%%%%%%%%%%%%%%%%%%%%%%%%
\section{Experiments}\label{sec:experiment}
%%%%%%%%%%%%%%%%%%%%%%%%%%%%%%%%%%%%%%%%%%%%%%%%%%%%%%%%%%%

\noindent\textbf{Implementation Details.} We use GNN as the point cloud representation extractor, where $K=24$ of KNN experimentally. The dimension $m=512$. The number of layers $\ell=5$. We train our network for 500 epochs with an initial learning rate of $1e^{-3}$. 
As our method is unsupervised, we fine-tune the model for 200 epochs with an initial learning rate of $1e^{-4}$ on the test split. 
We use the Adam optimizer and the experiment is performed in PyTorch 1.0.0 with a batch size of 26.

\noindent\textbf{Dataset and Evaluation Metric.} For synthetic dataset ModelNet40 \cite{wu_modelnet40_cvpr_2015}, following \cite{wang_dcp_iccv_2019,wang_prnet_nips_2019}, the rotation and translation along each axis between the \textit{source} and \textit{target} are in $[0^\circ, 45^\circ]$, $[-0.5, 0.5]$. For real dataset 7Scenes \cite{jamie_7scenes_cvpr_2013}, following \cite{huang_featuremetric_cvpr_2020}, the rotation is initialized in $[0^\circ, 60^\circ]$ and the translation is initialized in $[0, 1.0]$ along one randomly selected axis. Root mean square error (RMSE) and mean absolute error (MAE) in Euler angle and translation vector, and mean error in $SE(3)$ \cite{romain_define_ijcv_2018} are used as evaluation metrics, notated as RMSE(R), MAE(R), RMSE(t), MAE(t), and ME(T) respectively.

\vspace{-0.3cm}
\subsection{Evaluation on ModelNet40}
Following \cite{wang_dcp_iccv_2019,wang_prnet_nips_2019}, three experiment settings are applied here:
\textbf{1) Unseen Point Clouds} (\textit{UPC}): All point clouds are divided into training and test sets with the official setting. 
\textbf{2) Unseen Categories} (\textit{UC}): The first 20 categories in ModelNet40 are used for training and the rest for test.
\textbf{3) Noisy Data} (\textit{ND}): Random Gaussian noise (\ie\ $\mathcal{N}(0,0.01)$, and the sampled noise out of the range of $[-0.05,0.05]$ will be clipped) is added to the \textit{UPC} setting.

\noindent\textbf{Consistent Point Clouds.} Following \cite{wang_dcp_iccv_2019}, the input point clouds are consistent here. Two versions of PRNet have been evaluated, \ie, PRNet selects half points as the key points \cite{wang_prnet_nips_2019} and PRNet* selects all points as the key points. Besides using distance-based feature notated as UPCR w/ Dis, we also test our method using SPFH \cite{rusu_fpfh_icra_2009}, PFH \cite{PFH_rusu_iros_2008} as pose-invariant feature notated as UPCR w/ SPFH and UPCR w/ PFH respectively.
From \tabref{Tab:registration}, for \textbf{\textit{UPC}}, in all unsupervised methods, UPCR achieves the best performance. Meanwhile, our unsupervised method even obtains better performance under many metrics than supervised methods. Especially, UPCR outperforms the correspondences-free method, PointNetLK by a large margin.
For \textbf{\textit{UC}}, in all unsupervised methods, UPCR significantly outperforms all baselines. Our method achieves higher accuracy under most metrics than those supervised methods.
For \textbf{\textit{ND}}, UPCR is superior to DCP-v2 but inferior to PRNet and PRNet* under RMSE(R) while achieving the best results under other metrics. Then, in all UPCR pipelines, UPCR w/ PFH achieves the most accurate results due to the informative pose-invariant feature.

\begin{table}[ht]
\centering
\vspace{-0.6em}
\caption{Evaluation in the consistent point clouds. Note: ``\checkmark'' indicates the unsupervised method. }
\vspace{-0.6em}
\setlength\tabcolsep{3pt}

\resizebox{\linewidth}{!}{%
\begin{tabular}{lrrrrrrr|rrrrrr|rrr}
\toprule
\multicolumn{1}{l}{\multirow{2}{*}{Methods}} &\multicolumn{1}{c}{}& \multicolumn{3}{c}{RMSE(R)$\ \downarrow$} & \multicolumn{3}{c|}{MAE(R)$\ \downarrow$} & \multicolumn{3}{c}{RMSE(t)\ ($\times 10^{-2}$)$\ \downarrow$} & \multicolumn{3}{c|}{MAE(t)\ ($\times 10^{-2}$)$\ \downarrow$}& \multicolumn{3}{c}{{ME(T)$\ \downarrow$}} \\

\cmidrule(lr){3-5} \cmidrule(lr){6-8} \cmidrule(lr){9-11} \cmidrule(lr){12-14} \cmidrule(lr){15-17}

\multicolumn{1}{c}{} &\multicolumn{1}{c}{}
& \multicolumn{1}{c}{UPC} & \multicolumn{1}{c}{UC} & \multicolumn{1}{c}{ND} 
& \multicolumn{1}{c}{UPC} & \multicolumn{1}{c}{UC} & \multicolumn{1}{c|}{ND} 
& \multicolumn{1}{c}{UPC} & \multicolumn{1}{c}{UC} & \multicolumn{1}{c}{ND} 
& \multicolumn{1}{c}{UPC} & \multicolumn{1}{c}{UC} & \multicolumn{1}{c|}{ND}
& \multicolumn{1}{c}{UPC} & \multicolumn{1}{c}{UC} & \multicolumn{1}{c}{ND} \\ 
\midrule
\textbf{PointNetLK}\cite{aoki_ptlk_cvpr_2019} & 
    & 13.751 & 15.901 & 15.692 & 3.893  & 4.032  & 3.992 
    & 1.990  & 2.612  & 2.396  & 0.445  & 0.621  & 0.564 
    & 15.217 & 16.937 & 16.716 \\
\textbf{DCP-v2}\cite{wang_dcp_iccv_2019} & 
    & \textbf{1.094} & 3.256 & 8.417 & 0.752 & 2.102 & 5.685 
    & 0.172 & 0.632  & {0.117} & 0.463 & 2.337 & 2.337 
    & \textbf{1.774} &4.647   & 10.140\\
\textbf{PRNet}\cite{wang_prnet_nips_2019} & 
    & 1.722 & {3.060} & \textbf{3.218} & 0.665 & 1.326 & 1.446 
    & 0.637 & 1.010 & 0.465  & 0.759 & 0.838  & 0.838 
    & 2.579 & 4.172 & 5.021 \\
\textbf{PRNet*}\cite{wang_prnet_nips_2019}& 
    & 2.090  & 3.720  & 3.292  & 0.894  & 1.543  & 1.449 
    & 1.098  & 1.313  & 1.077  & 0.815  & 0.996  & 0.815 
    & 2.765  & 4.113  & 4.974  \\ %\arrayrulecolor{gray} \cdashline{1-14}[1.2pt/2pt]
%\cmidrule(r){1-14}
\textbf{ICP\cite{besl_icp_pami_1992}} & \checkmark
    & 12.282 & 12.707 & 11.971 & 4.613  & 5.075  & 4.497 
    & 4.774  & 4.853  & 4.832  & 0.228  & 0.236  & 0.234 
    & 16.719 & 16.911 & 13.276 \\
\textbf{FGR}\cite{zhou_fgr_eccv_2016} & \checkmark
    & 20.054 & 21.323 & 18.359 & 7.146  & 8.077  & 6.367 
    & 4.412  & 4.578  & 3.910  & 1.642  & 0.181  & 1.449 
    & 24.126 & 25.037 & 22.452 \\
{\textbf{ICP}+\textbf{PFH}\cite{PFH_rusu_iros_2008}}   & \checkmark
    & 7.713 & 10.287  & 8.751 & 1.143  & 2.017  & 2.376 
    & 1.143  & 1.798  & 1.004  & 0.115  & 0.107  & 0.193 
    & 8.872 & 12.035  & 11.284 \\
{\textbf{ICP}+\textbf{SPFH}\cite{rusu_fpfh_icra_2009}}   & \checkmark
    & 8.039 & 10.977  & 9.082 & 1.437  & 2.391  & 2.753 
    & 1.797  & 1.848   & 2.034  & 0.202  & 0.194  & 0.223 
    & 9.986 & 14.425  & 14.028 \\
\cmidrule(r){1-17}
\textbf{UPCR w/ Dis} &\checkmark 
    & 2.774  & 3.212  & 5.420  & {0.623} & {1.053} & {1.432} 
    & {0.023} & {0.201} & 0.151 & {0.005} & {0.061} & {0.105} 
    & 3.524  & 3.986  & 6.107\\
{\textbf{UPCR w/ SPFH}} &\checkmark 
    & 2.563  & 3.091  & 4.975  & {0.584} & {0.921} & {1.224} 
    & {0.020} & {0.188} & 0.134 & \textbf{0.004} & {0.052} & {0.100}   & 3.233   & 3.569 & 5.547\\
{\textbf{UPCR w/ PFH}} &\checkmark 
    & 2.373  & \textbf{3.011}  & 4.624  & \textbf{0.514} & \textbf{0.823} & \textbf{1.043} 
    & \textbf{0.018} & \textbf{0.156} & \textbf{0.112} & \textbf{0.004} & \textbf{0.040} & \textbf{0.087}   & 3.011   & \textbf{3.164} & \textbf{4.927}\\
\bottomrule
\end{tabular}%
}
\label{Tab:registration}
\vspace{-0.2cm}
\end{table}

\noindent\textbf{Partial-to-Partial.} Our method is robust to outliers because the disturbance in the pose-invariant part has been filtered out. Following the common protocol of PRNet \cite{wang_prnet_nips_2019}, we simulate partial scans of $\mathbf{X}$ and $\mathbf{Y}$ by randomly placing a point in 3D space and selecting its 768 nearest neighbors from the consistent point clouds respectively, where there are 1024 points in each original consistent point cloud. 
The performance is reported in \tabref{Tab:registration_partial}.
For \textbf{\textit{UPC}}, our UPCR achieves the best performance in all unsupervised methods. Besides, we obtain comparable or better results compared with the state-of-the-art PRNet. 
For \textbf{\textit{UC}}, UPCR achieves the best performance under all metrics, even outperforms PRNet. 
For \textbf{\textit{ND}}, our method achieves the best performance under most evaluation metrics over all unsupervised and supervised methods.
These results verify the robustness of our method to partial-to-partial cases. Besides, UPCR w/ PFH achieves the best performance among all UPCR pipelines because PFH feature is more informative.

\begin{table}[ht]
\centering
\vspace{-0.1cm}
\caption{Evaluation in the partial-to-partial setting. }
\vspace{-0.6em}
\setlength\tabcolsep{3pt}

\resizebox{\linewidth}{!}{%
\begin{tabular}{lrrrrrrr|rrrrrr|rrr}
\toprule
\multicolumn{1}{l}{\multirow{2}{*}{Methods}} &\multicolumn{1}{c}{}& \multicolumn{3}{c}{RMSE(R)$\ \downarrow$} & \multicolumn{3}{c|}{MAE(R)$\ \downarrow$} & \multicolumn{3}{c}{RMSE(t)$\ \downarrow$} & \multicolumn{3}{c|}{MAE(t)$\ \downarrow$} & \multicolumn{3}{c}{ME(T)$\ \downarrow$} \\
\cmidrule(lr){3-5} \cmidrule(lr){6-8} \cmidrule(lr){9-11} \cmidrule(lr){12-14} \cmidrule(lr){15-17}

\multicolumn{1}{c}{}&\multicolumn{1}{c}{} 
& \multicolumn{1}{c}{UPC} & \multicolumn{1}{c}{UC} & \multicolumn{1}{c}{ND} & \multicolumn{1}{c}{UPC} & \multicolumn{1}{c}{UC} & \multicolumn{1}{c|}{ND} & \multicolumn{1}{c}{UPC} & \multicolumn{1}{c}{UC} & \multicolumn{1}{c}{ND} & \multicolumn{1}{c}{UPC} & \multicolumn{1}{c}{UC} & \multicolumn{1}{c|}{ND} & \multicolumn{1}{c}{UPC} & \multicolumn{1}{c}{UC} & \multicolumn{1}{c}{ND} \\ 

\midrule
\textbf{PointNetLK}\cite{aoki_ptlk_cvpr_2019} & 
    & 16.735 & 22.943 & 19.939 & 7.550  & 9.655  & 9.076 
    & 0.045  & 0.061  & 0.057  & 0.025  & 0.033  & 0.032 
    &18.264  & 23.417 & 21.547 \\
\textbf{DCP-v2}\cite{wang_dcp_iccv_2019} & 
    & 6.709  & 9.769  & 6.883  & 4.448  & 6.954  & 4.534 
    & 0.027  & 0.034  & 0.028  & 0.020  & 0.025  & 0.021 
    & 7.847  & 10.527 & 7.912 \\
\textbf{PRNet}\cite{wang_prnet_nips_2019} & 
    & \textbf{3.199} & 4.986 & \textbf{4.323} & {1.454} & 2.329 & 2.051 
    & 0.016          & 0.021 & {0.017} & 0.010 & 0.015 & 0.012 
    & 4.428  & 5.746 & \textbf{5.013}\\ 
%\arrayrulecolor{gray} \cdashline{1-14}[1.2pt/2pt] 
\textbf{ICP}\cite{besl_icp_pami_1992}   & \checkmark
    & 33.683 & 34.894 & 35.067 & 25.045 & 25.455 & 25.564 
    & 0.293  & 0.293  & 0.294  & 0.250  & 0.251  & 0.250 
    & 36.745 & 38.294 & 40.012 \\
\textbf{FGR}\cite{zhou_fgr_eccv_2016}   & \checkmark
    & 11.238 & 9.932  & 27.635 & 2.832  & 1.952  & 13.794 
    & 0.030  & 0.038  & 0.070  & 0.008  & 0.007  & 0.039 
    & 13.243 & 12.476 & 30.127 \\
{\textbf{ICP}+\textbf{PFH}\cite{PFH_rusu_iros_2008}}   & \checkmark
    & 14.389 & 10.743  & 27.976 & 3.578  & 4.217  & 14.278 
    & 0.147  & 0.172  & 0.144  & 0.108  & 0.117  & 0.193 
    & 17.013 & 12.896 & 30.104 \\
{\textbf{ICP}+\textbf{SPFH}\cite{rusu_fpfh_icra_2009}}   & \checkmark
    & 18.647 & 24.312  & 29.904 & 4.015  & 4.993  & 15.673 
    & 0.190  & 0.201  & 0.159  & 0.207  & 0.198  & 0.214 
    & 20.778 & 27.746 & 32.368 \\
\cmidrule(r){1-17}
\textbf{UPCR w/ Dis} & \checkmark
    & 3.901 & {3.713} & 5.522 & 1.518 & {1.138} & {1.702} & {0.009} & {0.008} & 0.020 & {0.004} & \textbf{0.002} & {0.010} 
    & 4.542 & 4.879 & 6.748 \\ 
{\textbf{UPCR w/ SPFH}} & \checkmark
    & 3.728 & {3.387} & 5.217 & 1.274 & {1.008} & {1.468} & \textbf{0.008} & {0.008} & 0.016 & {0.004} & \textbf{0.002} & \textbf{0.008}
    & 4.034 & 4.374 & 6.038 \\ 
{\textbf{UPCR w/ PFH}} & \checkmark
    & 3.271 & \textbf{3.014} & 4.876 & \textbf{1.012} & \textbf{0.971} & \textbf{1.195} & \textbf{0.008} & \textbf{0.007} & \textbf{0.011} & \textbf{0.003} & \textbf{0.002} & \textbf{0.008} 
    & \textbf{3.512} & \textbf{4.014} & 5.429 \\ 
    \bottomrule
\end{tabular}%
}
\label{Tab:registration_partial}
\vspace{-0.6cm}
\end{table}

%%%%%%%%%%%%%%%%%%%%%%%%%%%%%%%%%%%%%%%%%%%
\subsection{Evaluation on 7Scenes} 
%%%%%%%%%%%%%%%%%%%%%%%%%%%%%%%%%%%%%%%%%%%
In \tabref{Tab:generalization}, we evaluate our UPCR on 7Scenes \cite{jamie_7scenes_cvpr_2013}, which is a remote scene RGB-D data including Heads, Office, Chess, Fire, Pumpkin, Redkitchen, Stairs scenes. We use the 3D point cloud information only. 
Since the real scene data is very challenging, state-of-the-art supervised point cloud registration methods, such as DCP-v2 and PRNet fail to estimate the transformation. For UPCR, we provide the model trained on ModelNet40, fine-tuned on Heads, fine-tuned on the corresponding test scene respectively. Results are consistent: 1) The model trained on ModelNet40 shows poor performance on scene data but is still better than baselines. 2) Models fine-tuned on Heads and corresponding test scenes show accurate and very similar performance, \ie the model fine-tuned on one scene can obtain accurate results on other scenes.

%%%%%%%%%%%%%%%%%%%%%%%%%%%%%%%%%%%%%%%%%%%%%%%%%
\vspace{-10pt}
\subsection{Ablation Study} \label{sec::ablation}
%%%%%%%%%%%%%%%%%%%%%%%%%%%%%%%%%%%%%%%%%%%%%%%%%
%%%%%%%%%%%%%%%%%%%%%%%%%%%%%%%%%%%%%%%%%%%%%%%%%%%%%%%%%%%%%%%
\noindent\textbf{1. Rotation Solver.}  \label{sec:::ablation_rotation}
%%%%%%%%%%%%%%%%%%%%%%%%%%%%%%%%%%%%%%%%%%%%%%%%%%%%%%%%%%%%%%%
There are several rotation representations, including rotation matrix, 6D expression \cite{Zhou_6D_CVPR_2019}, quaternion, Euler angle, \etc, which are compared in \tabref{Tab:ablation_rotation_solver}. The Euler angle representation achieves the best performance. We reckon this is because the network can converge more easily to achieve better results in a more compact representation.
% the network can achieve better results with less DoF, which makes it converge easily.

\begin{table}[h]
	\renewcommand\arraystretch{1.0}
	\vspace{-0.1cm}
	\caption{Performance of UPCR w/ Dis method in different rotation representation solvers on consistent point clouds under \textit{UPC}.}	
	\vspace{-0.4cm}
	\setlength\tabcolsep{3pt}
	\begin{center}
		\resizebox{\linewidth}{!}{
			\begin{tabular}{lcccccc}
				\toprule
				{Methods}&Variables&RMSE(R)&MAE(R)& \makecell[c]{RMSE(t)($\times 10^{-2}$)} &\makecell[c]{MAE(t)($\times 10^{-2}$)}&{ME(T)}\\
				\midrule				
				\multicolumn{1}{l}{\textbf{Rotation Matrix}}     &9
				&\multicolumn{1}{r}{34.894}      &\multicolumn{1}{r}{25.455}   &\multicolumn{1}{c}{0.293}       &\multicolumn{1}{c}{0.025} &37.976\\
				
				\multicolumn{1}{l}{\textbf{6D expression}}       &6
				&\multicolumn{1}{r}{7.972}       &\multicolumn{1}{r}{3.810}    &\multicolumn{1}{c}{0.076}       &\multicolumn{1}{c}{0.036} &10.286\\
				
				\multicolumn{1}{l}{\textbf{Quaternion}}          &4
				&\multicolumn{1}{r}{3.502}      &\multicolumn{1}{r}{\textbf{0.524}}    &\multicolumn{1}{c}{0.038}       &\multicolumn{1}{c}{\textbf{0.005}} &4.743\\
				
				\multicolumn{1}{l}{\textbf{Euler Angle}}         &3
				&\multicolumn{1}{r}{\textbf{2.774}}       &\multicolumn{1}{r}{0.623}    &\multicolumn{1}{c}{\textbf{0.023}}       &\multicolumn{1}{c}{\textbf{0.005}} &\textbf{3.524}\\
				\toprule
		\end{tabular}}
		\label{Tab:ablation_rotation_solver}
	\end{center}
	\vspace{-0.4cm}
\end{table}

\begin{figure}[h]
    \centerline{\includegraphics[width=0.99\linewidth]{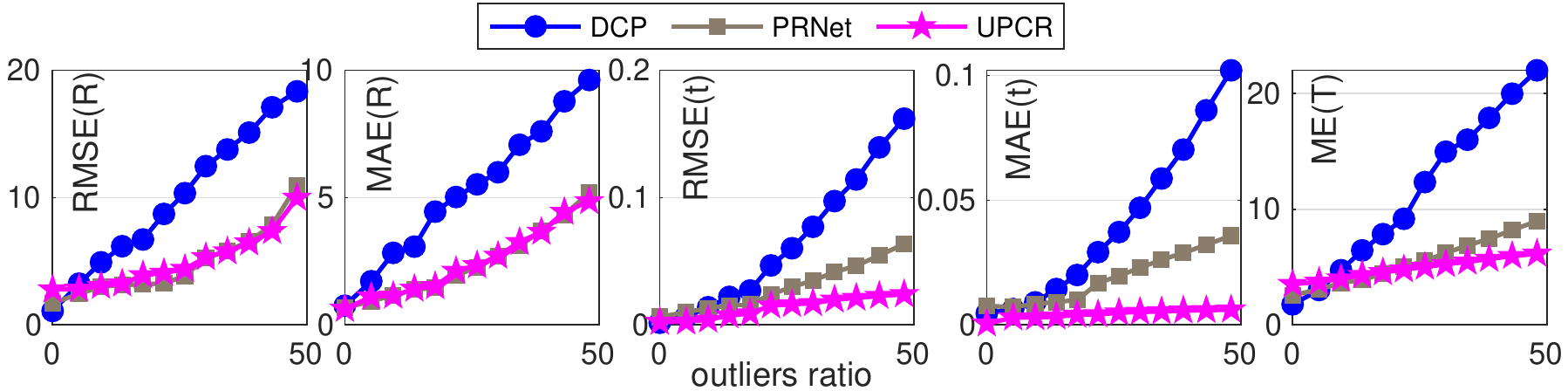}}
    \vspace{-0.2cm}
	\caption{Performance decreases as the outliers ratio (\%) increases.} 
	\label{Fig:degeneration}
    \vspace{-0.6cm}
\end{figure}

\noindent\textbf{2. Performance vs. Outliers Ratios.} \figref{Fig:degeneration} shows the performance degeneration as outliers ratio increases. 
% Our \textbf{unsupervised} method shows comparable performance to PRNet in rotation estimation and better translation estimation.
Compared to PRNet, our \textbf{unsupervised} UPCR w/ Dis method has comparable rotation estimation and better translation estimation results. Totally speaking, when outliers ratio increases, our method is more robust than these supervised methods.

%%%%%%%%%%%%%%%%%%%%%%%%%%%%%%%%%%%%%%%%%%%%%%%%%%%%%%%%%%%%%%%
\noindent\textbf{3. Completeness of Pose-Invariant Feature.}
%%%%%%%%%%%%%%%%%%%%%%%%%%%%%%%%%%%%%%%%%%%%%%%%%%%%%%%%%%%%%%%
We can observe that UPCR with more informative pose-invariant features, \eg SPFH, PFH achieves better performance than using distance-based feature since SPFH and PFH represent the pose-invariant part more completely. Here, to verified this conclusion further, 
we employ seven pose-invariant features, \ie \textbf{distance-based feature}, \textbf{PPF} \cite{ppf-foldnet_deng_eccv_2018}, \textbf{SPFH} \cite{rusu_fpfh_icra_2009}, \textbf{PFH} \cite{PFH_rusu_iros_2008}, \textbf{distance-based + PPF}, \textbf{distance-based + SPFH} and \textbf{distance-based + PPF + SPFH} to extract the pose-invariant representation variants, respectively. Then, the global representation, and the seven pose-related representations returned by ``subtracting'' these variants are used to perform an ablation experiment.

\begin{figure}[ht]
\vspace{-0.4cm}
	\centerline{\includegraphics[width=240pt,height=80pt]{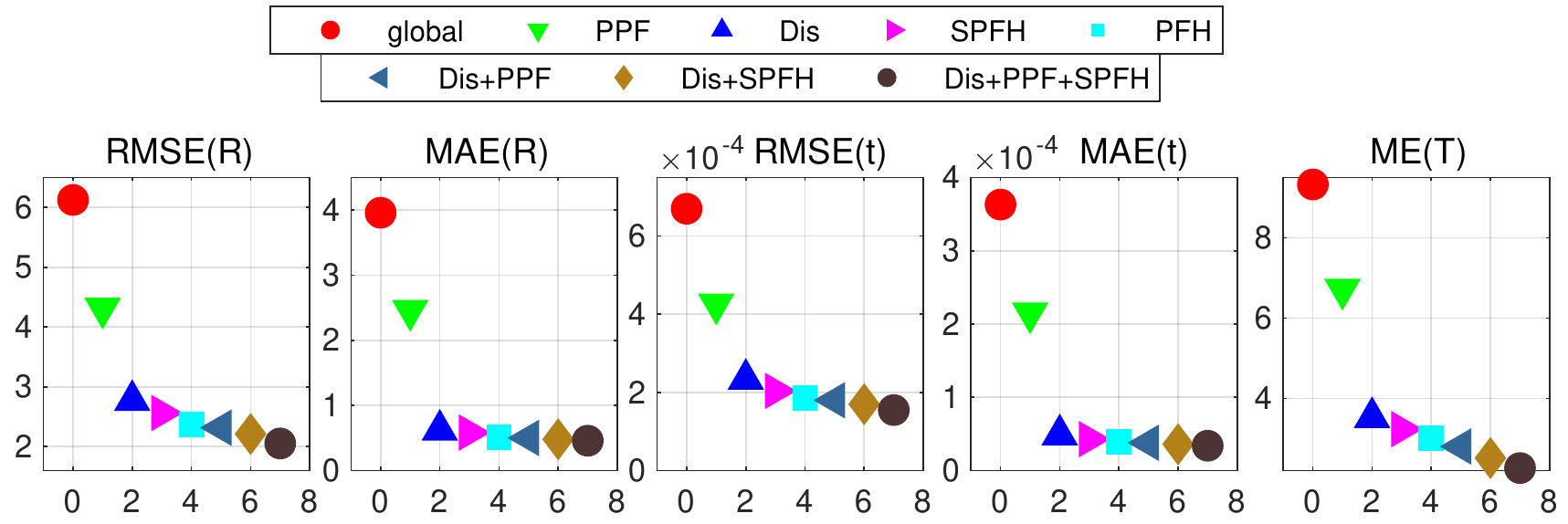}}
	\vspace{-0.2cm}
	\caption{Incremental experiment of pose-invariant feature.} 
	\label{Fig:invariable_feature}
\vspace{-0.3cm}
\end{figure}

From \figref{Fig:invariable_feature}, regressing pose via the entire point cloud underperforms purified pose-related part representation. The performance of using distance-based + PPF + SPFH is better than using one or two of them. Based on these observations, we conclude that our distance-based feature cannot represent the pose-invariant part completely, and some pose-invariant information still exists in pose-related representation obtained via designed entropy subtraction. Moreover, more informative pose-invariant feature, which results in more purified pose-related representation, helps obtain better registration performance. Although our distance-based feature underperforms several typical features, it has achieved comparable or better performance than the state-of-the-art supervised methods.

%%%%%%%%%%%%%%%%%%%%%%%%%%%%%%%%%%%%%%%%%%%%%%%%%%%%%%%%%%%%%%%%%%%%%%%%%%%%%%%%%%%%
\noindent\textbf{4. Time Complexity.}  \label{sec:::ablation_time}
%%%%%%%%%%%%%%%%%%%%%%%%%%%%%%%%%%%%%%%%%%%%%%%%%%%%%%%%%%%%%%%%%%%%%%%%%%%%%%%%%%%
We report the average inference time of each input sample with 1024 points on GTX 1080 Ti in \tabref{Tab:times}. Our UPCR w/ Dis achieves advanced time-efficiency due to the simple distance-based feature. UPCR w/ PFH shows poor time-efficiency due to the complex feature extraction.

\begin{table}[ht]
	\renewcommand\arraystretch{1.0}
	\vspace{-0.1cm}
	\caption{Inference time on point cloud with 1024 points.}
	\setlength\tabcolsep{3pt}
	\vspace{-0.4cm}
	\begin{center}
		\resizebox{\linewidth}{!}{
			\begin{tabular}{lcccccccc}
				\toprule
				%\multirow{2}{*}{\textbf{Methods}}&\multicolumn{2}{c}{\textbf{Rotation}}&\multicolumn{2}{|c}{\textbf{Translation}} \\
				%\cline{2-5}
				Methods & PointNetLK & DCP & PRNet & PRNet* & UPCR w/ Dis & UPCR w/ SPFH & UCPR w/ PFH\\
				\midrule				
				% Time(s)&0.06632&0.01098&0.04548&0.04722&\textbf{0.00972}\\
				Time(ms) & 66.32 & 10.98 & 45.48 & 47.22 & \textbf{9.71} & 2140.53 & 8134.34\\
				\bottomrule
		\end{tabular}}
		\label{Tab:times}
	\end{center}
	\vspace{-0.45cm}
\end{table}

\begin{table*}[!h]
\centering
% \vspace{-0.6em}
\caption{Evaluation on 7Scenes. \texttt{ModelNet40} indicates that the model is trained on ModelNet40, \texttt{Heads} means that the model is fine-tuned on Heads, \texttt{test scene} means that the model is fine-tuned on the corresponding test scene. }
\vspace{-0.6em}
\setlength\tabcolsep{3pt}
\resizebox{\linewidth}{!}{%
\begin{tabular}{lccc|ccc|ccc|ccc|ccc|ccc|ccc}
\toprule
\multicolumn{1}{l}{\multirow{2}{*}{Methods}} & \multicolumn{3}{c}{Heads} 
                                             & \multicolumn{3}{c}{Office}
                                             & \multicolumn{3}{c}{Chess} 
                                             & \multicolumn{3}{c}{Fire}
                                             & \multicolumn{3}{c}{Pumpkin}
                                             & \multicolumn{3}{c}{Redkitchen}
                                             & \multicolumn{3}{c}{Stairs} \\
\cmidrule(lr){2-4} \cmidrule(lr){5-7} \cmidrule(lr){8-10} \cmidrule(lr){11-13} \cmidrule(lr){14-16} \cmidrule(lr){17-19} \cmidrule(lr){20-22}

\multicolumn{1}{c}{} & \multicolumn{1}{c}{RMSE(R)} & \multicolumn{1}{c}{MAE(R)} & \multicolumn{1}{c}{ME(T)}
                     & \multicolumn{1}{c}{RMSE(R)} & \multicolumn{1}{c}{MAE(R)} & \multicolumn{1}{c}{ME(T)} 
                     & \multicolumn{1}{c}{RMSE(R)} & \multicolumn{1}{c}{MAE(R)} & \multicolumn{1}{c}{ME(T)}
                     & \multicolumn{1}{c}{RMSE(R)} & \multicolumn{1}{c}{MAE(R)} & \multicolumn{1}{c}{ME(T)}
                     & \multicolumn{1}{c}{RMSE(R)} & \multicolumn{1}{c}{MAE(R)} & \multicolumn{1}{c}{ME(T)}
                     & \multicolumn{1}{c}{RMSE(R)} & \multicolumn{1}{c}{MAE(R)} & \multicolumn{1}{c}{ME(T)}
                     & \multicolumn{1}{c}{RMSE(R)} & \multicolumn{1}{c}{MAE(R)} & \multicolumn{1}{c}{ME(T)}\\

\midrule
\textbf{DCP-v2}\cite{wang_dcp_iccv_2019} & 79.448 & 53.914 & 87.646
                                         & 77.268 & 49.594 & 85.476
                                         & 71.298 & 52.921 & 80.374
                                         & 75.264 & 50.924 & 83.714
                                         & 70.853 & 44.758 & 77.983
                                         & 79.023 & 51.043 & 86.142
                                         & 75.849 & 53.269 & 84.347\\
                                         
\textbf{PRNet}\cite{wang_prnet_nips_2019}& 75.719 & 49.808 & 84.276
                                         & 72.164 & 48.623 & 78.146
                                         & 65.972 & 47.026 & 71.419
                                         & 70.215 & 43.267 & 77.842
                                         & 66.473 & 42.125 & 73.982
                                         & 74.285 & 45.681 & 80.276
                                         & 70.364 & 47.231 & 77.451\\ 
\cmidrule(r){1-22}
\textbf{UPCR w/ Dis}(\texttt{ModelNet40})& 23.614 & 17.823 & 25.946
                                         & 18.503 & 14.308 & 21.047
                                         & 20.693 & 16.232 & 22.384
                                         & 19.735 & 15.467 & 22.046
                                         & 20.266 & 15.738 & 23.451
                                         & 19.651 & 15.254 & 21.137
                                         & 22.195 & 17.477 & 24.156\\
                                         
\textbf{UPCR w/ Dis}(\texttt{Heads})     & 1.424  & 0.913  & {2.184}
                                         & 0.997  & 0.689  & 1.478
                                         & 2.576  & 1.678  & 3.947
                                         & 1.093  & 0.682  & 2.024
                                         & 1.161  & 0.818  & 2.454
                                         & 0.967  & 0.641  & 1.989
                                         & 1.291  & 0.866  & 3.014\\
\textbf{UPCR w/ Dis}(\texttt{test scene})& 1.422  & 0.910  & 2.220
                                         & 0.998  & 0.690  & 1.466
                                         & 2.577  & 1.679  & 3.920
                                         & 1.093  & 0.681  & 2.019
                                         & 1.164  & 0.834  & 2.462
                                         & 0.953  & 0.653  & 2.011
                                         & 1.227  & 0.894  &2.917\\

\cdashline{1-22}[2.2pt/1.2pt]
\textbf{UPCR w/ SPFH}(\texttt{ModelNet40})& 20.141 & 15.436 & 23.778
                                         & 17.498 & 13.411 & 19.766
                                         & 18.455 & 14.112 & 19.978
                                         & 16.433 & 13.477 & 18.479
                                         & 18.472 & 14.010 & 20.147
                                         & 17.496 & 14.788 & 19.022
                                         & 18.452 & 15.787 & 20.196\\
                                         
\textbf{UPCR w/ SPFH}(\texttt{Heads})    & 1.221  & 0.900  & 2.102
                                         & 0.879  & 0.654  & 1.244
                                         & 2.334  & 1.433  & 3.748
                                         & 1.007  & 0.641  & 1.997
                                         & 1.024  & 0.776  & 2.101
                                         & 0.756  & 0.502  & 1.701
                                         & 1.204  & \textbf{0.745}  & 2.456\\
\textbf{UPCR w/ SPFH}(\texttt{test scene})& 1.207 & 0.801  & 2.003
                                         & 0.842  & 0.614  & 1.213
                                         & 2.374  & 1.441  & 3.696
                                         & \textbf{1.004}  & 0.593  & 1.867
                                         & 0.944  & 0.798  & 2.114
                                         & 0.806  & 0.602  & 1.879
                                         & 1.007  & 0.787  & 2.532\\
\cdashline{1-22}[2.2pt/1.2pt]
\textbf{UPCR w/ PFH}(\texttt{ModelNet40})& 19.011 & 14.217 & 20.179
                                         & 17.244 & 13.207 & 19.463
                                         & 18.017 & 13.877 & 19.062
                                         & 16.014 & 13.038 & 17.844
                                         & 18.027 & 13.545 & 19.233
                                         & 16.314 & 14.022 & 18.131
                                         & 17.032 & 13.977 & 17.936\\
                                         
\textbf{UPCR w/ PFH}(\texttt{Heads})    & 1.200  & 0.900  & 2.004
                                         & 0.843  & 0.642  & 1.204
                                         & 2.301  & 1.377  & 3.644
                                         & 1.007  & 0.601  & 1.827
                                         & 1.001  & 0.731  & \textbf{1.927}
                                         & \textbf{0.711}  & \textbf{0.422}  & \textbf{1.535}
                                         & 1.007  & \textbf{0.745}  & \textbf{2.421}\\
\textbf{UPCR w/ PFH}(\texttt{test scene})& \textbf{1.113} & \textbf{0.724}   & \textbf{1.775}
                                         & \textbf{0.802}  & \textbf{0.543}  & \textbf{1.007}
                                         & \textbf{2.075}  & \textbf{1.214}  & \textbf{3.499}
                                         & \textbf{1.004}  & \textbf{0.544}  & \textbf{1.752}
                                         & \textbf{0.902}  & \textbf{0.756}  & 2.021
                                         & 0.788  & 0.566  & 1.789
                                         & \textbf{1.001}  & 0.746  & 2.455\\
                                   
\midrule
\multicolumn{1}{c}{} & \multicolumn{1}{c}{RMSE(t)} & \multicolumn{1}{c}{MAE(t)} & \multicolumn{1}{c}{ME(T)}
                     & \multicolumn{1}{c}{RMSE(t)} & \multicolumn{1}{c}{MAE(t)} & \multicolumn{1}{c}{ME(T)}
                     & \multicolumn{1}{c}{RMSE(t)} & \multicolumn{1}{c}{MAE(t)} & \multicolumn{1}{c}{ME(T)}
                     & \multicolumn{1}{c}{RMSE(t)} & \multicolumn{1}{c}{MAE(t)} & \multicolumn{1}{c}{ME(T)}
                     & \multicolumn{1}{c}{RMSE(t)} & \multicolumn{1}{c}{MAE(t)} & \multicolumn{1}{c}{ME(T)}
                     & \multicolumn{1}{c}{RMSE(t)} & \multicolumn{1}{c}{MAE(t)} & \multicolumn{1}{c}{ME(T)}
                     & \multicolumn{1}{c}{RMSE(t)} & \multicolumn{1}{c}{MAE(t)} & \multicolumn{1}{c}{ME(T)} \\
                     
\midrule
\textbf{DCP-v2}\cite{wang_dcp_iccv_2019} & 0.185 & 0.123 &-
                                         & 0.176 & 0.105 &-
                                         & 0.192 & 0.116 & -
                                         & 0.201 & 0.114 & -
                                         & 0.193 & 0.113 & -
                                         & 0.179 & 0.122 & -
                                         & 0.202 & 0.137 & -\\
                                         
\textbf{PRNet}\cite{wang_prnet_nips_2019}& 0.080 & 0.054 & -
                                         & 0.077 & 0.062 &-
                                         & 0.093 & 0.071 &-
                                         & 0.088 & 0.063 &-
                                         & 0.079 & 0.068 &-
                                         & 0.083 & 0.065 &-
                                         & 0.079 & 0.059 &- \\
                                         
\cmidrule(r){1-22}
\textbf{UPCR w/ Dis}(\texttt{ModelNet40})                & 0.044 & 0.031 &-
                                         & 0.048 & 0.032 & -
                                         & 0.051 & 0.038 &-
                                         & 0.047 & 0.035 & -
                                         & 0.061 & 0.045 & -
                                         & 0.048 & 0.036 & -
                                         & 0.054 & 0.041 &-\\
                                         
\textbf{UPCR w/ Dis}(\texttt{Heads})     & 0.026 & 0.018 &-
                                         & 0.023 & 0.018  &-
                                         & 0.030 & 0.021 &-
                                         & 0.019 & 0.014 &-
                                         & 0.029 & 0.019 &-
                                         & 0.020 & 0.014 &-
                                         & 0.023 & 0.015 &-\\
                                         
\textbf{UPCR w/ Dis}(\texttt{test scene})& 0.025 & 0.020 &-
                                         & 0.021 & \textbf{0.015} &-
                                         & 0.033 & 0.027 &-
                                         & 0.022 & 0.017 &-
                                         & 0.024 & 0.021 &-
                                         & 0.022 & 0.013 &-
                                         & 0.021 & 0.012 &-\\
\cdashline{1-22}[2.2pt/1.2pt]
\textbf{UPCR w/ SPFH}(\texttt{ModelNet40})& 0.032 & 0.027 &-
                                         & 0.040 & 0.026 & -
                                         & 0.041 & 0.032 &-
                                         & 0.042 & 0.030 & -
                                         & 0.055 & 0.041 & -
                                         & 0.042 & 0.032 & -
                                         & 0.047 & 0.036 &-\\
                                         
\textbf{UPCR w/ SPFH}(\texttt{Heads})    & 0.023 & 0.016 &-
                                         & 0.021 & 0.017  &-
                                         & \textbf{0.027} & \textbf{0.018} &-
                                         & 0.017 & 0.011 &-
                                         & 0.021 & \textbf{0.015} &-
                                         & 0.017 & 0.011 &-
                                         & 0.019 & 0.012 &-\\
                                         
\textbf{UPCR w/ SPFH}(\texttt{test scene}) & 0.023 & 0.017 &-
                                         & 0.020 & \textbf{0.015} &-
                                         & 0.030 & 0.025 &-
                                         & 0.019 & 0.015 &-
                                         & 0.020 & 0.020 &-
                                         & 0.020 & 0.011 &-
                                         & \textbf{0.018} & \textbf{0.010} &-\\
\cdashline{1-22}[2.2pt/1.2pt]
\textbf{UPCR w/ PFH}(\texttt{ModelNet40})& 0.030 & 0.024 &-
                                         & 0.035 & 0.025 & -
                                         & 0.037 & 0.030 &-
                                         & 0.034 & 0.030 & -
                                         & 0.050 & 0.040 & -
                                         & 0.042 & 0.030 & -
                                         & 0.043 & 0.034 &-\\
                                         
\textbf{UPCR w/ PFH}(\texttt{Heads})    & 0.023 & \textbf{0.015} &-
                                         & 0.020 & 0.017  &-
                                         & 0.025 & \textbf{0.018} &-
                                         & \textbf{0.016} & \textbf{0.010} &-
                                         & 0.020 & \textbf{0.015} &-
                                         & \textbf{0.016} & \textbf{0.010} &-
                                         & 0.019 & 0.011 &-\\
                                         
\textbf{UPCR w/ PFH}(\texttt{test scene}) & \textbf{0.022} & 0.016 &-
                                         & \textbf{0.018} & \textbf{0.015} &-
                                         & \textbf{0.027} & 0.022 &-
                                         & 0.017 & 0.015 &-
                                         & \textbf{0.018} & 0.017 &-
                                         & 0.019 & \textbf{0.010} &-
                                         & \textbf{0.018} & \textbf{0.010} &-\\
\bottomrule
\end{tabular}%
}
\label{Tab:generalization}
\vspace{-0.3cm}
\end{table*}

%%%%%%%%%%%%%%%%%%%%%%%%%%%%%%%%%%%%%%%%%%%%%%%%%%%%%%%%%%%%
\section{Conclusion} \label{sec:conclusion}
%%%%%%%%%%%%%%%%%%%%%%%%%%%%%%%%%%%%%%%%%%%%%%%%%%%%%%%%%%%%

In this paper, we have proposed a correspondence-free unsupervised 3D point cloud registration method from the representation separation perspective. Our proposed method not only filters out the disturbance in pose-invariant representation but also is robust to partial-to-partial or noisy point clouds. Experiments on benchmark datasets demonstrate the superiority and robustness of our unsupervised 3D point cloud registration method, which achieves comparable if not better performance than the state-of-the-art supervised methods.

{\small
%\balance
% \bibliographystyle{ieee_fullname}
% \bibliography{Unsupervised_Registration_Ref}

% \bibliographystyle{IEEEtran}
\bibliographystyle{IEEEtranS}
\bibliography{references}
}

\end{document}